%% file: main.tex
\documentclass{article}

\usepackage{microtype}
\usepackage{graphicx}
\usepackage{booktabs} 
\usepackage{amsmath,amssymb,amsfonts}
\usepackage{bm}
\usepackage{bbm}
\usepackage{caption}
\usepackage{subfig}

\newcommand{\be}{\begin{equation}}
\newcommand{\ee}{\end{equation}}
\newcommand{\ba}{\begin{align}}
\newcommand{\ea}{\end{align}}

\usepackage{hyperref}

\usepackage[accepted]{icml2021}

\icmltitlerunning{Federated Active Learning (F-AL): an Efficient Annotation Strategy for Federated Learning}

\begin{document}

\onecolumn
\icmltitle{Federated Active Learning (F-AL):  an Efficient Annotation Strategy for Federated Learning}





\begin{icmlauthorlist}
\icmlauthor{Jin-Hyun Ahn}{MGH}
\icmlauthor{Kyungsang Kim}{MGH}
\icmlauthor{Jeongwan Koh}{MGH}
\icmlauthor{Quanzheng Li}{MGH}
\end{icmlauthorlist}

\icmlaffiliation{MGH}{Massachusetts General Hospital and Harvard Medical School, Massachusetts, USA}

\icmlcorrespondingauthor{Quanzheng Li}{li.quanzheng@mgh.harvard.edu}

\icmlkeywords{Federated Learning, Active Learning}

\vskip 0.3in



\printAffiliationsAndNotice{}  

\begin{abstract}
Federated learning (FL) has been intensively investigated in terms of communication efficiency, privacy, and fairness. However, efficient annotation, which is a pain point in real-world FL applications, is less studied. In this project, we propose to apply active learning (AL) and sampling strategy into the FL framework to reduce the annotation workload. We expect that the AL and FL can improve the performance of each other complementarily. In our proposed federated active learning (F-AL) method, the clients collaboratively implement the AL to obtain the instances which are considered as informative to FL in a distributed optimization manner. We compare the test accuracies of the global FL models using the conventional random sampling strategy, client-level separate AL (S-AL), and the proposed F-AL. We empirically demonstrate that the F-AL outperforms baseline methods in image classification tasks.   
\end{abstract}


\input{a}

\input{b}

\input{c}
\input{d}
\input{e}
\input{f}



\bibliography{main}
\bibliographystyle{icml2021}

\end{document}

%% file: a.tex
\section{Introduction}

Federated learning (FL) \cite{mcmahan2017communication} enables the collaborative training from datasets residing on distributed clients with the help of a parameter server. In numerous previous works, including \cite{smith2017federated,li2020federated,sattler2019robust,li2019convergence}, the superiority of FL has been validated through numerical results and convergence analysis in independently identically distributed (IID) and non-IID datasets. While the recent literature related to FL primarily addresses communication efficiency, fairness, robustness, privacy of FL, and personalization, almost all of the previous works have assumed that the training datasets at clients are perfectly ready to be used for training. 

However, the annotation step should not be overlooked or ignored for the practical implementations of FL, like other machine learning (ML), since the cost for labeling is generally high and might be even dominant over the FL itself. Considering this problem, we study the annotation strategies in the FL framework, where the clients participating in FL should label their datasets prior to FL execution. For the annotations, we apply active learning (AL) \cite{settles2009active} at each client participating in FL. Because labeling all the instances is rarely a practical or cost-effective, AL aims to maximize the model's performance based on the fewest samples by selectively sampling and labeling the most informative instances.

To validate the proposed method, we establish a FL framework with the annotation step, where various active learning strategies in the FL are compared: 1) conventional FL with random sampling, 2) client-level separated active learning (S-AL), and 3) the proposed federated active learning (F-AL). In the F-AL, the clients collaboratively execute the AL to select the instances that are considered informative to FL in a distributed optimization manner. For the S-AL and F-AL, the state-of-the-art AL algorithms are incorporated.

The AL certainly outperforms the random sampling in the centralized learning. However, to the best of the authors’ knowledge, there has been no work for considering the AL in the FL framework and investigating the effect of AL on the performance of FL. This work demonstrates that AL can surprisingly reduce the cost of labeling for FL, and the cost-saving is fascinating in the FL environments. Furthermore, we show that the proposed F-AL considerably improves the performance of AL in the FL environment. We summarize our contributions below:

\begin{itemize}
    \item We establish a general FL framework combining with the annotation step. We evaluate the three types of methods: conventional FL with random sampling, S-AL, and F-AL. With the S-AL, the clients independently apply AL in their datasets. The F-AL encourages the clients' collaboration for AL.
    
    \item We empirically demonstrate that the AL is effective in the FL environment through various experiments with AL algorithms and datasets. The numerical result indicates that the AL methods outperform random sampling in terms of test accuracy of global FL models.
    
    \item 
    We demonstrate that F-AL outperforms the other methods. We highlight that the F-AL magnifies the benefit of AL in the FL environment.  
    
\end{itemize}

%% file: b.tex
\section{Related Work \& Background}

\subsection{Federated Learning}
FL can be categorized into cross-device FL and cross-silo FL \cite{kairouz2019advances}. In both of FL, data is locally generated and stored while the data is centrally managed and distributed to clients in the setting of datacenter distributed learning. The cross-device FL supposes that the clients are an enormous number of mobile or IoT devices connected by wi-fi or slow connections. Therefore, uplink communication is the main bottleneck of performance. Furthermore, it generally encounters fresh training samples which are never seen before since most clients participate only once in an entire FL process.

On the other hand, cross-silo FL typically supposes that the distribution scale is $2$-$100$ clients, which are generally different organizations or geo-distributed data centers such as hospitals or banks. Therefore,  it supposes that all clients are available during the whole FL process, and the clients' datasets are repeatedly used for training from round to round. The performance degradation due to communication bottleneck is not as severe as the case of cross-device FL. Instead, the performance heavily depends on the number or quality of the training dataset \cite{fenza2021data}.  

FL can be executed in various ways in terms of optimization strategy of the knowledge among the clients. The most classic algorithms in FL are federated stochastic gradient descent (FedSGD), or federated averaging (FedAvg) \cite{mcmahan2017communication} which are based on the averaging of the clients' parameters. Beyond the vanilla algorithms, FedProx \cite{li2020federated} and FedDF \cite{lin2020ensemble} tackles the systems and statistical heterogeneity, FedMA \cite{wang2020federated} and FetchSGD \cite{rothchild2020fetchsgd} alleviate the communication bottleneck, and TERM \cite{li2020tilted} and Ditto \cite{li2021ditto} are related to the fairness and robustness in personalized FL.

\subsection{Active Learning}
AL selects the informative instances to be labeled prior to the other instances and aims to maximize the model's performance based on the fewest samples. It has been demonstrated that AL can considerably reduce the number of labeling samples and alleviate the heavy burden of cost for annotation \cite{settles2009active,ren2021survey}. In fact, it has been proved that an effective AL strategy can theoretically obtain exponential acceleration in the efficiency of labeling \cite{balcan2009agnostic}. Even when it is applied in the area of deep learning (DL), the cost saving in the annotation is much more fascinating since DL has its explicit limitation due to the high cost of labeling the numerous instances, even brutal in the professional field that requires rich knowledge \cite{bengio2007greedy,krizhevsky2012imagenet}.

The sampling strategies of AL can be categorized into uncertainty-based sampling, representation-based sampling, other sampling strategies leverage the characteristic of deep learning such as learning loss (LL) \cite{yoo2019learning}, Monte-Carlo dropout (MC-dropout) \cite{gal2016dropout}, adversarial active learning \cite{sinha2019variational,kim2021task}, and hybrid sampling using the strategies jointly. Uncertainty-based sampling \cite{lewis1994sequential,beluch2018power} queries the instances which are the most uncertain to the model trained on the current training samples. Representation-based sampling \cite{geifman2017deep,sener2017active} measures the representativeness of unlabeled samples and encourages the sampling strategy to select the instances from different areas of the distribution. Since the sampling strategy only concerned with uncertainty may skew the model due to the similarity of the sampled instances in a particular distribution, the balance between uncertainty and representativeness is one of the main issues in the performance of AL strategies \cite{sener2017active}.

Furthermore, most of the recent work related to AL focuses on the AL strategy for DL by leveraging the aspects of ML model such as estimated training loss \cite{yoo2019learning}, length of gradient \cite{freytag2014selecting} and MC dropout \cite{gal2016dropout} for uncertainty estimation. Adversarial active learning \cite{sinha2019variational,wang2020dual,zhang2020state,kim2021task} trains a generative adversarial network (GAN) structured auxiliary network which learns a low dimensional latent space and discriminates the labeled and unlabeled samples in order to select unlabeled instances which are most different from the labeled instances. Furthermore, \cite{cho2021mcdal} recently proposed Maximum Classifier Discrepancy for Active Learning (MCDAL) which is the first work that leverages classifier discrepancy for sampling in active learning.

%% file: c.tex
\section{Problem Definition}\label{sec:problem}
This section provides FL framework where the annotation step is included before the execution of FL. We first introduce the FL environment comprising a parameter server and clients. The annotation step in the FL framework is described and formulated in more detail. Furthermore, we provide the AL which the clients execute in the annotation step.

\subsection{Federated Learning Environment}
We consider a cross-silo FL comprising a parameter server and $M$ clients. The clients store their own local dataset $\mathcal{U}_{m}$, $m=1,\dots,M$ which are the unlabeled datasets. Before the start of FL, each $m$-th client selects the instances from $\mathcal{U}_{m}$ and labels the instances to obtain $\mathcal{D}_{m}=\left\{\bm{x}_{i},\bm{y}_{i}\right\}$ where $\mathcal{L}_{m}=\left\{\bm{x}_{i}\right\}$ is the selected instance from $\mathcal{U}_{m}$ and $\bm{y}_{i}$ is the label of $\bm{x}_{i}$. We denote the sampling function as $\mathcal{A}\left(\cdot\right)$ and the selected instances, $\mathcal{L}_{m}$, as 
\be\label{sampling}
\mathcal{L}_{m}=\mathcal{A}\left(\mathcal{U}_{m}\right),
\ee
for $m=1,\dots,M$.

Let $\bm{\theta} \in \mathbb{R}^{D}$ denote the global model to be optimized in FL. The local loss $F_{m} \left( \bm{\theta}\right)$ at the $m$-th client is $F_{m} \left( \bm{\theta}\right)= \frac{1}{\left|\mathcal{D}_{m} \right|} \sum_{\bm{u} \in \mathcal{D}_{m}} f \left( \bm{\theta}, \bm{u} \right)$,
where $\mathcal{D}_{m}$ is the labeled dataset at the $m$-th client and $f\left(\cdot\right)$ is the loss function determined by the network model. Accordingly, the global loss $F\left(\bm{\theta}\right)$ is defined as $
F\left(\bm{\theta}\right)=\frac{1}{ \left|\bigcup_{m=1}^{M} \mathcal{D}_{m} \right|} \sum_{\bm{u} \in \bigcup_{m=1}^{M} \mathcal{D}_{m}} f \left( \bm{\theta}, \bm{u} \right)$.
The goal of FL is to train the optimized parameter $\bm{\theta}^{*}$ minimizing the global loss, namely
$\bm{\theta}^{*}=\underset{\theta}{\mathrm{argmin}} ~ F\left(\bm{\theta}\right)$.

The FedSGD \cite{mcmahan2017communication} is applied for the FL updates, where $\bm{\theta}^{*}$ is obtained through iterative stochastic gradient descent (SGD) allowing the parallel computation of gradients at the clients. The parameter vector $\bm{\theta}_{t}$ at the $t$-th iteration is updated according to $\bm{\theta}_{t+1} = \bm{\theta}_{t}  - \eta_{t}  \sum _{m=1}^{M} \frac{n_{m}}{n}\bm{g}_{m }\left(\bm{\theta}_{t}\right)$,
where $\eta_{t}$ is the learning rate at the $t$-th iteration, $n=\sum_{m=1}^{M} n_{m}$, $n_{m}=\left|\mathcal{D}_{m} \right|$  and $\bm{g}_{m }\left(\bm{\theta}_{t}\right) \in \mathbb{R}^{D}$ is the stochastic gradient of $\bm{\theta}_{t}$ computed at the $m$-th client as 
$\bm{g}_{m}\left(\bm{\theta}_{t}\right)=\frac{1}{\left|\mathcal{D}_{m}\right|} \sum_{\bm{u} \in \mathcal{D}_{m}}  \nabla f\left(\bm{\theta}_{t},\bm{u}\right)$. The update is equivalently given by 
\be\label{average}
\bm{\theta}_{t+1} = \sum_{m=1}^{M}\frac{n_{m}}{n} \bm{\theta}^{m}_{t+1},
\ee
where
\be\label{update}
\bm{\theta}^{m}_{t+1} =  \bm{\theta}_{t} - \eta_{t} \bm{g}_{m }\left(\bm{\theta}_{t}\right).
\ee
In this work, the FedAvg \cite{mcmahan2017communication} computes the converged solution at each client by repeating \eqref{update} multiple times before the average. The overall FL framework is summarized in Algorithm \ref{alg:FedAvg}.

\begin{figure}[ht]
\vskip 0.2in
\begin{center}
\centerline{\includegraphics[width=0.5\columnwidth]{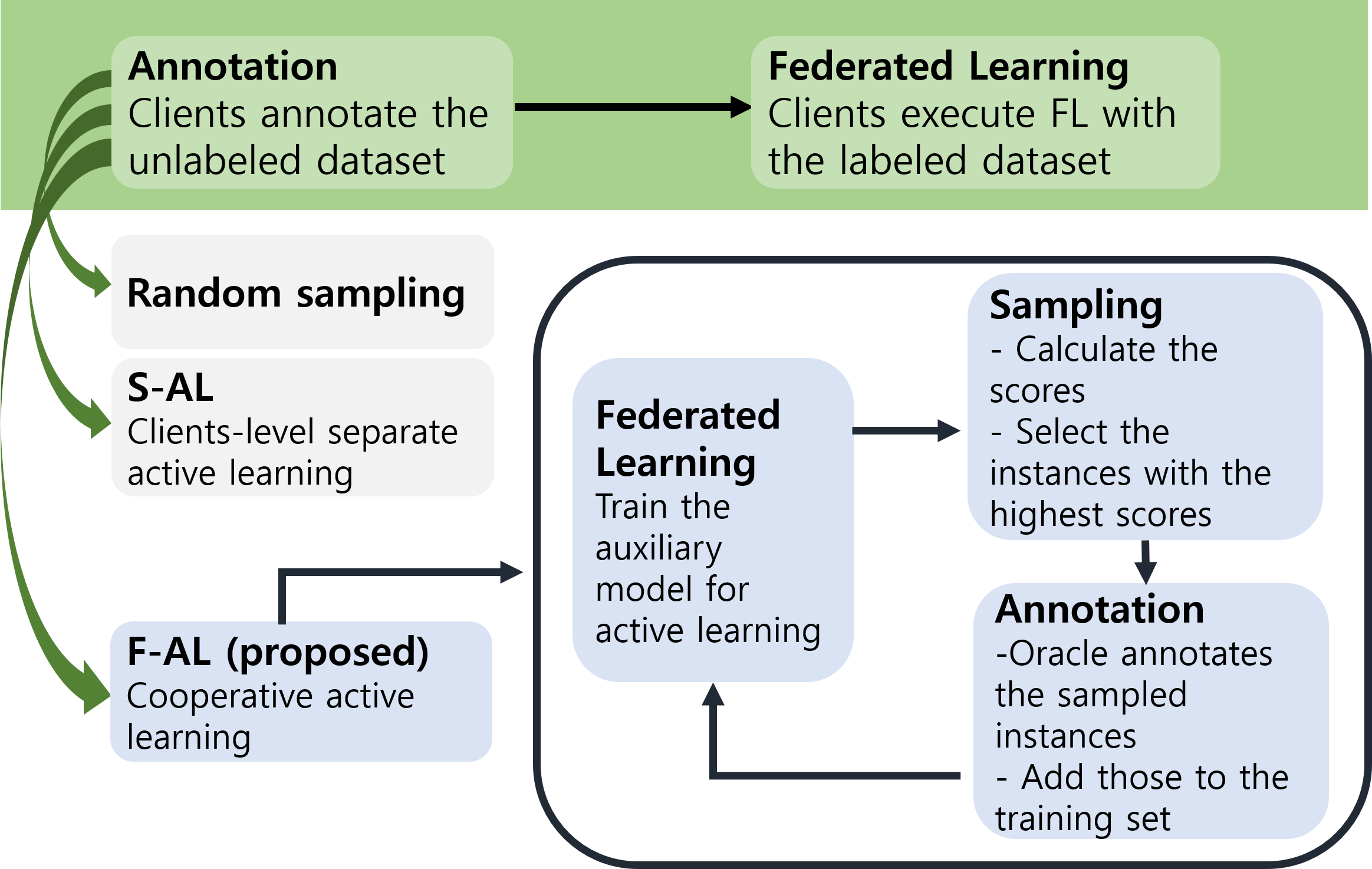}}
\caption{Annotation strategies for federated learning.}
\label{overview}
\end{center}
\vskip -0.2in
\end{figure}

\begin{algorithm}[tb]
   \caption{Federated Learning with annotation step}
   \label{alg:FedAvg}
\begin{algorithmic}
   \STATE {\bfseries Input:} unlabeled datasets, $\left\{\mathcal{U}_{m}\right\}_{m=1}^{M}$ 
   \STATE \phantom{{\bfseries Input:}} initialized model, $\bm{\theta}_{1}$
   \STATE \phantom{{\bfseries Input:}} learning rate, $\left\{\eta_{t}\right\}_{t}$
   \begin{ALC@g}
   \STATE {\bfseries Annotation step:}
   \begin{ALC@g}
   \FOR{$m=1$ {\bfseries to} $M$}
   \STATE annotate $\mathcal{L}_{m}=\mathcal{A}\left(\mathcal{U}_{m}\right)$ to obtain $\mathcal{D}_{m}$
   \ENDFOR
   \end{ALC@g}
   \STATE {\bfseries FL step:}
   \begin{ALC@g}
   \FOR{each round $t=1,2,\dots$ }
   \STATE {\bfseries Client executes:}
   \begin{ALC@g}
   \STATE do multiple iterations of \eqref{update}
   \end{ALC@g} 
   \STATE {\bfseries Server executes:}
   \begin{ALC@g}
   \STATE average model parameters as in \eqref{average}
   \end{ALC@g}
   \ENDFOR
   \end{ALC@g} 
   \STATE {\bfseries return} $\texttt{FedAvg}\left(\left\{ \mathcal{D}_{m} \right\}_{m}^{M} \;\middle|\; \bm{\theta}_{1} ,\left\{\eta_{t}\right\}_{t} \right)$
   \end{ALC@g}
\end{algorithmic}
\end{algorithm}

\subsection{Active Learning}\label{sec:problem:AL}
In the proposed FL framework, we introduce the sampling function, $\mathcal{A}\left(\cdot\right)$, which finds the instances to be labeled from the unlabeled dataset prior to the process of FL. For an example of random sampling, the acquired sample instances from the unlabeled dataset $\mathcal{U}$ is
$\mathcal{A}\left(\mathcal{U}\right)=\texttt{random}(\mathcal{U}, b)$,
where $\texttt{random}\left(\mathcal{U}, b\right)$ is to randomly choose the $b$ instances from $\mathcal{U}$. In terms of the sampling function, the goal of AL is to find the best sampling function which selects the most informative and effective instances to the performance of the main task. 

Most of the AL algorithms generally searches instances with the highest score in the unlabeled data pool \cite{mccallumzy1998employing} as
\be
\mathcal{A}\left(\mathcal{U}\right)= \underset{\mathcal{L} \subseteq \mathcal{U},~\left| \mathcal{L} \right|=b, ~ x \in \mathcal{L}}{\mathrm{argmax}} ~ S\left(x\right),
\ee
where $b$ is the budget of sampling, and $S\left(x\right)$ is the score function of $x$. The score function of effective AL should perfectly reflect the potential informativeness of instances in the unlabeled dataset. Hence, the AL algorithms can be described by how to design $S\left(\cdot\right)$. The score includes uncertainty, representativeness such as diversity, density, training loss, and dissimilarity to the labeled dataset.

Since the informativeness depends on the current labeled dataset, the score function is also conditioned on the current state of the labeled dataset.
Hence, the score is generally calculated based on the trained model with the current labeled dataset, namely
\be\label{scorefunction}
S\left(x\right)=S\left(x\; \middle| \; \mathcal{D}\right)=S\left(x\; \middle| \; \bm{\phi} \left(\mathcal{D}\right) \right),
\ee
where $\bm{\phi} \left(\mathcal{D}\right)$ is the auxiliary model that is trained with the labeled dataset $\mathcal{D}$, starting from the randomly initialized model. Furthermore, AL adopts multiple rounds for sampling and gradually samples from the unlabeled dataset. When it is desired to add $b$ instances to be labeled after $K$ rounds, it samples $b/K$ instances at each round. We summarize the description of AL algorithm, $\mathcal{A}\left(\cdot\right)$, in Algorithm 2.

\begin{algorithm}[tb]
   \caption{Active Learning, $\mathcal{A}\left(\cdot\right)$}
   \label{alg:AL}
\begin{algorithmic}
   \STATE {\bfseries Input:} unlabeled dataset, $\mathcal{U}^{1}$
   \STATE \phantom{{\bfseries Input:}} initially labeled dataset, $\mathcal{D}^{1}$
   \STATE \phantom{{\bfseries Input:}} number of AL round, $K$
   \STATE \phantom{{\bfseries Input:}} initialized models for $\bm{\phi}$, $\{\bm{\phi}^{k}\}_{k=1}^{K}$
   \STATE \phantom{{\bfseries Input:}} number of annotation budget, $b$
   \begin{ALC@g}
   \FOR{$k=1$ {\bfseries to} $K$}
   \STATE train $\bm{\phi}\left(\mathcal{D}^{k}\right)$, starting from $\bm{\phi}^{k}$
   \STATE sample $\mathcal{L}^{k}$,
   \be\label{samplinginAL}
   \mathcal{L}^{k}=\underset{\mathcal{L} \subseteq \mathcal{U}^{k},~\left| \mathcal{L} \right|=\frac{b}{K}, ~ x \in \mathcal{L}}{\mathrm{argmax}} ~ S\left(x\; \middle| \; \bm{\phi}\left(\mathcal{D}^{k}\right) \right) 
   \ee
   \STATE $\hat{\mathcal{D}}^{k}=\texttt{annotate}\left(\mathcal{L}^{k}\right)$
   \STATE $\mathcal{D}^{k+1 }   = \mathcal{D}^{k} \cup \hat{\mathcal{D}}^{k}$
   \STATE $\mathcal{U}^{k+1 } = \mathcal{U}^{k} - \mathcal{L}^{k}$
   \ENDFOR
   \STATE {\bfseries return} $\mathcal{D}^{K+1}$ with size of $\left|\mathcal{D}^{1}\right|+b$
   \end{ALC@g}
\end{algorithmic}
\end{algorithm}

\begin{algorithm}[tb]
   \caption{Federated Active Learning, $\mathcal{A}\left(\cdot\right)$}
   \label{alg:FAL}
\begin{algorithmic}
   \STATE {\bfseries Input:} unlabeled dataset, $\left\{\mathcal{U}^{1}_{m}\right\}_{m=1}^{M}$
   \STATE \phantom{{\bfseries Input:}} initially labeled dataset, $\left\{\mathcal{D}^{1}_{m}\right\}_{m=1}^{M}$
   \STATE \phantom{{\bfseries Input:}} number of AL round, $K$
   \STATE \phantom{{\bfseries Input:}} initialized models for $\bm{\phi}$, $\{\bm{\phi}^{k}\}_{k=1}^{K}$
   \STATE \phantom{{\bfseries Input:}} number of annotation budget, $\left\{b_{m}\right\}_{m=1}^{M}$
   \begin{ALC@g}
   \FOR{$k=1$ {\bfseries to} $K$}
   \STATE {\bfseries FL step:}
   \begin{ALC@g}
   \STATE train $\bm{\phi}^{k}_{FL}$, starting from $\bm{\phi}^{k}$
   \be
\bm{\phi}_{FL}^{k}= \texttt{FedAvg}\left(\left\{ \mathcal{D}_{m}^{k} \right\}_{m=1}^{M} \;\middle|\; \bm{\phi}^{k} ,\left\{\eta^{k}_{t}\right\}_{t} \right) 
\ee
   \end{ALC@g}
   \STATE {\bfseries Sampling step:}
   \begin{ALC@g}
   \FOR{$m=1$ {\bfseries to} $M$}
   \STATE sample $\mathcal{L}^{k}_{m}$,
   \be\label{samplinginFAL}
   \mathcal{L}^{k}_{m}=\underset{\mathcal{L} \subseteq \mathcal{U}^{k}_{m},~\left| \mathcal{L} \right|=\frac{b_{m}}{K}, ~ x \in \mathcal{L}}{\mathrm{argmax}} ~ S\left(x\; \middle| \; \bm{\phi}^{k}_{FL}\right) 
   \ee
   \STATE $\hat{\mathcal{D}}^{k}_{m}=\texttt{annotate}\left(\mathcal{L}_{m}^{k}\right)$
   \STATE $\mathcal{D}_{m}^{k+1 }   = \mathcal{D}_{m}^{k} \cup \hat{\mathcal{D}}_{m}^{k}$
   \STATE $\mathcal{U}_{m}^{k+1 } = \mathcal{U}_{m}^{k} - \mathcal{L}_{m}^{k}$
   \ENDFOR
   \end{ALC@g}
   \ENDFOR
   \STATE {\bfseries return} $\mathcal{D}_{m}^{K+1}$ with size of $\left|\mathcal{D}_{m}^{1}\right|+b_{m}$, $m=1,\dots,M$
   \end{ALC@g}
\end{algorithmic}
\end{algorithm} 

%% file: d.tex
\section{Federated Active Learning (F-AL)}
This section introduces AL methods in the FL framework: S-AL and F-AL. In the benchmark scheme of conventional FL adopting random sampling, we set $\mathcal{A}\left(\mathcal{U}\right)=\texttt{random}\left(\mathcal{U},b\right)$ in the Algorithm 1, as we introduce in the Section \ref{sec:problem:AL}.

\subsection{Separate Active Learning (S-AL)}
In S-AL, the clients separately perform the AL before the FL execution. With S-AL, the $m$-th client applies $\mathcal{A}\left(\cdot\right)$ of Algorithm \ref{alg:AL} to its unlabeled dataset at the annotation step in the FL framework. The S-AL directly leverages the AL in the FL framework, including the annotation step, and might seem straightforward. However, no related work establishes AL in FL and investigates the effect of AL in FL. 

\subsection{Federated Active Learning (F-AL)}
In S-AL, the clients independently accomplish AL and achieve the instances which are informative to the local datasets as in $\eqref{samplinginAL}$. At the $k$-th round, the $m$-th client selects $x$ with the highest score, $S\left(x\; \middle| \; \bm{\phi}\left(\mathcal{D}_{m}^{k}\right) \right)$, where $\mathcal{D}_{m}^{k}$ denotes $\mathcal{D}^{k}$ in the Algorithm \ref{alg:AL} in the perspective of $m$-th client.

Since the clients execute FL after the annotation step, however, it should be the main objective to obtain instances which are informative to the aggregate labeled dataset, $\mathcal{D}^{k}_{total}=\bigcup_{m=1}^{M} \mathcal{D}_{m}^{k}$ as in \eqref{scorefunction}. Therefore, the score function in F-AL is conditioned on $\mathcal{D}^{k}_{total}$ and defined as $S\left(x\; \middle| \; \bm{\phi}\left(\mathcal{D}_{total}^{k}\right) \right)$. But, $\bm{\phi}\left(\mathcal{D}_{total}^{k}\right)$ cannot be built because $\mathcal{D}_{m}^{k}$, $m=1,\dots,M$ should not be compiled to satisfy the constraint of FL. Thus, we replace $\bm{\phi}\left(\mathcal{D}_{total}^{k}\right)$ with the model trained by FL, $\bm{\phi}_{FL}^{k}$, which is  
\be\label{FALmodel}
\bm{\phi}_{FL}^{k}= \texttt{FedAvg}\left(\left\{ \mathcal{D}_{m}^{k} \right\}_{m=1}^{M} \;\middle|\; \bm{\phi}^{k} ,\left\{\eta^{k}_{t}\right\}_{t} \right). 
\ee
Accordingly, with F-AL, clients carry out FL to obtain the score function that represents the informativeness of the aggregate labeled dataset.

As a more clear perspective for the explanation, uncertainty \cite{lewis1994sequential,beluch2018power} can illustrate the ground why $\bm{\phi}_{FL}^{k}$ should be leveraged for the calculation of score function in order to improve FL performance. If the AL applies uncertainty-based sampling or the sampling related to uncertainty, referred as to task aware AL in \cite{kim2021task}, it utilizes the uncertainty score, which is measured by the main task model trained with the current labeled dataset. Therefore, the auxiliary model is the main task model, namely, $\bm{\phi}\left(\mathcal{D}^{k}\right)=\bm{\theta}\left(\mathcal{D}^{k}\right)$ in Algorithm \ref{alg:AL} or the set of auxiliary models includes the main tasks model in the case of several auxiliary models. Hence, we remark that the auxiliary model should also be obtained through FL since the main task model is trained by FL.

After attaining $\bm{\phi}_{FL}^{k}$ in F-AL, the instances can be ideally sampled as 
\be \label{samplingideal}
   \mathcal{L}^{k}=\underset{\mathcal{L} \subseteq \mathcal{U}^{k},~\left| \mathcal{L} \right|=\frac{b}{K}, ~ x \in \mathcal{L}}{\mathrm{argmax}} ~ S\left(x\; \middle| \; \bm{\phi}_{FL}^{k}  \right), 
\ee
where $\mathcal{U}^{k}= \bigcup_{m=1}^{M} \mathcal{U}^{k}_{m}$, $b=\sum_{m=1}^{M} b_{m}$. Under the annotation workload condition that the $m$-th client annotates $b_{m}/K$ instances at each round, we have $\mathcal{L}^{k}= \bigcup_{m=1}^{M} \mathcal{L}^{k}_{m}$, where
\be\label{samplinginFAL2}
\mathcal{L}^{k}_{m}=\underset{\mathcal{L} \subseteq \mathcal{U}^{k}_{m},~\left| \mathcal{L} \right|=\frac{b_{m}}{K}, ~ x \in \mathcal{L}}{\mathrm{argmax}} ~ S\left(x\; \middle| \; \bm{\phi}^{k}_{FL}\right). 
\ee
Therefore, each $m$-th client samples $\mathcal{L}_{m}^{k}$ as in \eqref{samplinginFAL2} and follows the remaining steps in Algorithm \ref{alg:AL}. In fact, the sampling step in $\eqref{samplingideal}$ can be executed at the server by exchanging the scores and indices of instances. However, we do not go any further since it might break the fairness of annotation workload among clients.  

%% file: e.tex
\section{Experiments}
This section provides the implementation details and the numerical results with related discussion. We compare the
performance of FL using the random sampling, S-AL, and the proposed F-AL in image classification tasks. The annotation strategies are applied for the annotation step in the Algorithm \ref{alg:FedAvg}, where the test accuracy of the obtained model is measured for the performance metric. For the image classification tasks, we evaluate the performances of the annotation strategies on the classical public datasets, Fashion-MNIST \cite{xiao2017fashion}, CIFAR-10 \cite{krizhevsky2009learning}, and CIFAR-100 \cite{krizhevsky2009learning}. The Fashion-MNIST dataset is a more challenging alternative dataset for the MNIST dataset. It consists of a training dataset of 60,000 images for 10 types of clothing and a test dataset of 10,000 images. CIFAR-10 and CIFAR-100 contain 50,000 training images and 10,000 test images. CIFAR-10 has 10 classes, while CIFAR-100 has 100 classes.

\subsection{Active learning algorithms}
First, we evaluate the performance of annotation strategies when the AL algorithm is the recently proposed Maximum Classifier Discrepancy for Active Learning (MCDAL) \cite{cho2021mcdal} which is one of the state-of-the-art AL algorithms. It utilizes the prediction discrepancies between two auxiliary classifiers after learning the auxiliary classifiers to maximize the discrepancies. It replaces the classic uncertainty with the discrepancies in the predictions of the auxiliary classifiers. It empirically demonstrates that this approach outperforms the state-of-the-art AL algorithms on the several
image classifications, including CIFAR-10 and CIFAR-100.

For more discussion, we evaluate the performance of annotation strategies for the various kinds of AL algorithms to achieve consistency in performance comparison. The first category of the AL algorithms uncertainty-related AL algorithms. This category of AL algorithms includes the classic uncertainty-based sampling with maximum entropy \cite{lewis1994sequential}, MC-dropout with maximum entropy \cite{gal2016dropout}, Learning Loss (LL) \cite{yoo2019learning}, and MCDAL \cite{cho2021mcdal}. The other AL algorithms are the core-set approach \cite{sener2017active} and variational adversarial active learning (VAAL). The core-set approach is the most widely used AL among the representative-based AL in the literature, and the VAAL represents the recent adversarial AL algorithms \cite{wang2020dual,zhang2020state,kim2021task}. All of the algorithms consider the main task model as the auxiliary model for AL. In the LL, MCDAL, and VAAL algorithms, the auxiliary models are additionally assumed and trained for AL. Therefore, the other models can be trained by FL in addition to the main task models for F-AL. For the evaluation of LL, however, we locally train the auxiliary model because no improvements are observed with the FL of auxiliary models in our experiments.

\subsection{Implementation details}
In the experiments, we assume that $M=5$ clients respectively have disjoint $10000$ images where $10 \%$ of the dataset is initially labeled. In our active learning setup, the $10 \%$ of the dataset is added to the labeled dataset at the sampling step of each round. We repeat this AL rounds until the total dataset is labeled. Hence, we set $b=10000$, $K=10$, and measure the test accuracy of FL model at each $k$-th round of AL, $\texttt{FedAvg}\left(\left\{ \mathcal{D}^{k}_{m} \right\}_{m}^{M} \;\middle|\; \bm{\theta}^{k}_{1} ,\left\{\eta_{t}^{k}\right\}_{t} \right)$.

We apply the Resnet-18 \cite{he2016deep} for the base architecture of main task model for all the exemplary tasks.
In the FL implementation, the main task models are optimized by SGD with the learning rate of $5 \times 10^{-2}$ and learning rate decay of $0.997$ per global iteration. The number of the local epoch is $1$, and the global iteration ends when the training loss at the clients decrease below thresholds, $1 \times 10 ^{-3}$, $5 \times 10 ^{-4}$, and $1.5 \times 10 ^{-3}$ for Fashion-MNIST, CIFAR-10, and CIFAR-100, respectively. In the independent learning for S-AL, we use SGD with a learning rate $1 \times 10^{-2}$ and step decay of $0.997$ at every epoch. Independent learning follows the same stopping criteria as FL. We use random horizontal flips for data augmentations. For the result of the experiments, we use the average accuracy of three runs.

\begin{figure}[t]
\vskip 0.2in
\begin{center}
\centerline{\includegraphics[width=0.4\columnwidth]{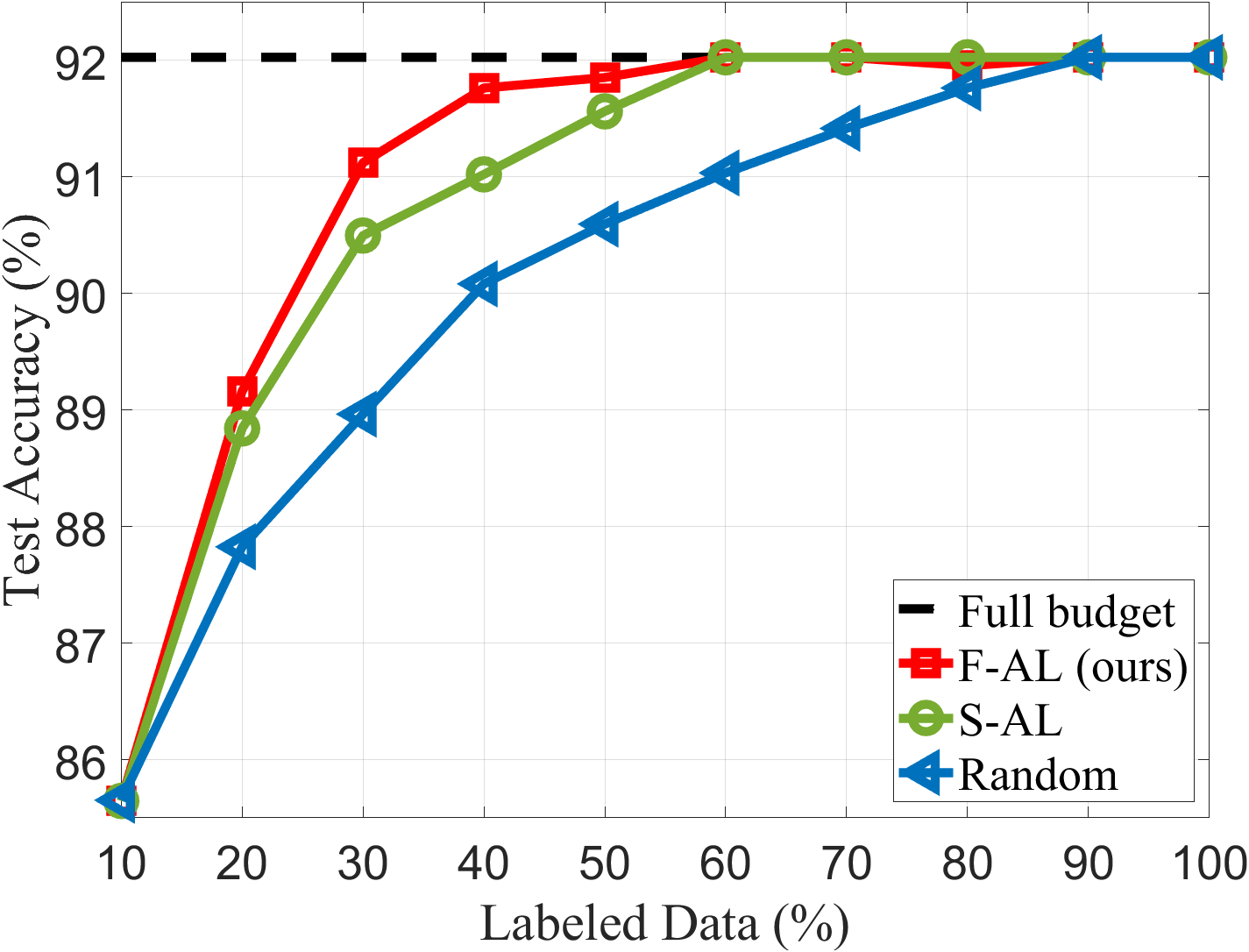}}
\caption{Test accuracies of global model trained by FL per rounds on Fashion-MNIST.}
\label{repre_mnist}
\end{center}
\vskip -0.2in
\end{figure}

\begin{figure}[t]
\vskip 0.2in
\begin{center}
\centerline{\includegraphics[width=0.4\columnwidth]{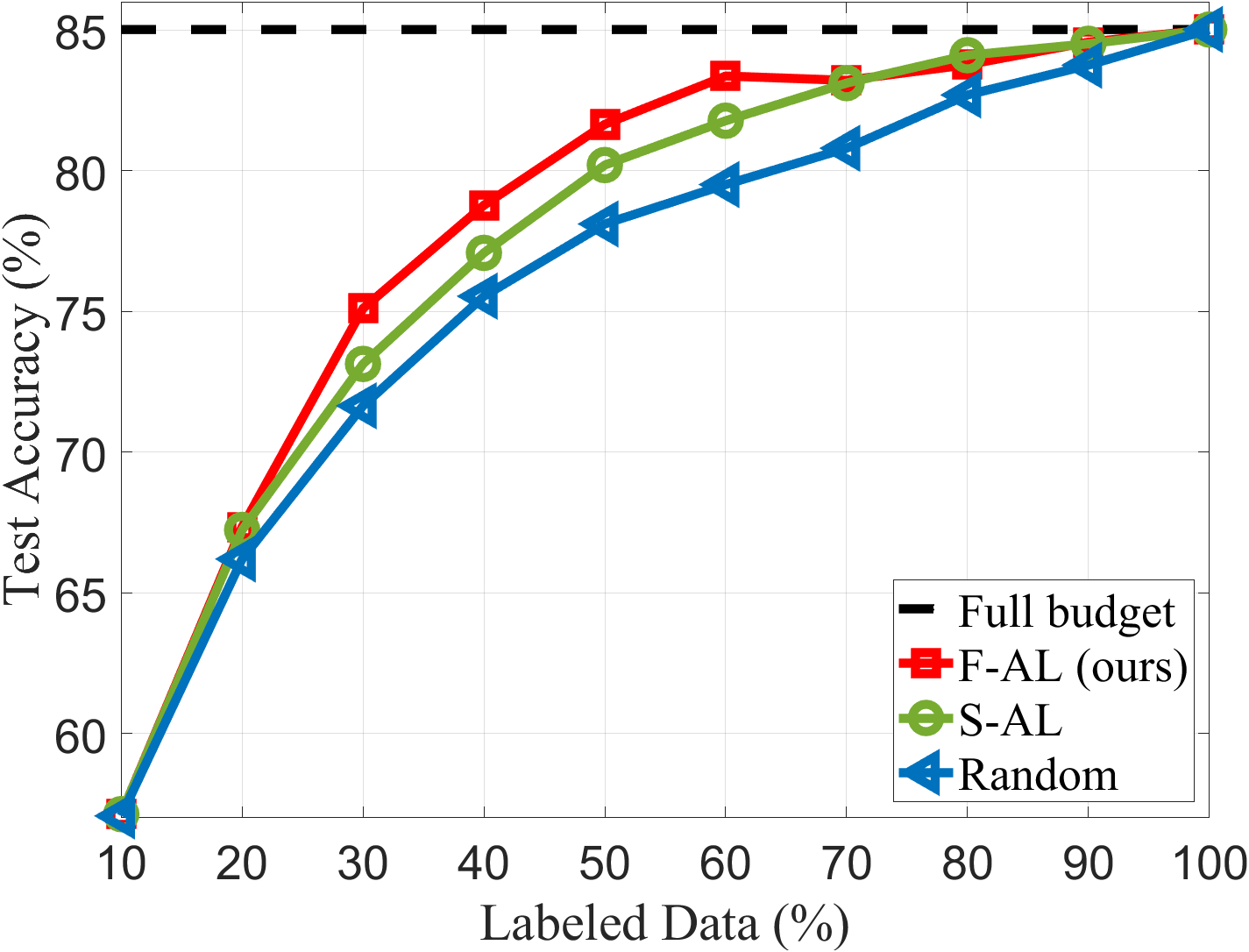}}
\caption{Test accuracies of global model trained by FL per rounds on CIFAR-10.}
\label{repre_cifar10}
\end{center}
\vskip -0.2in
\end{figure}

\begin{figure}[t]
\vskip 0.2in
\begin{center}
\centerline{\includegraphics[width=0.4\columnwidth]{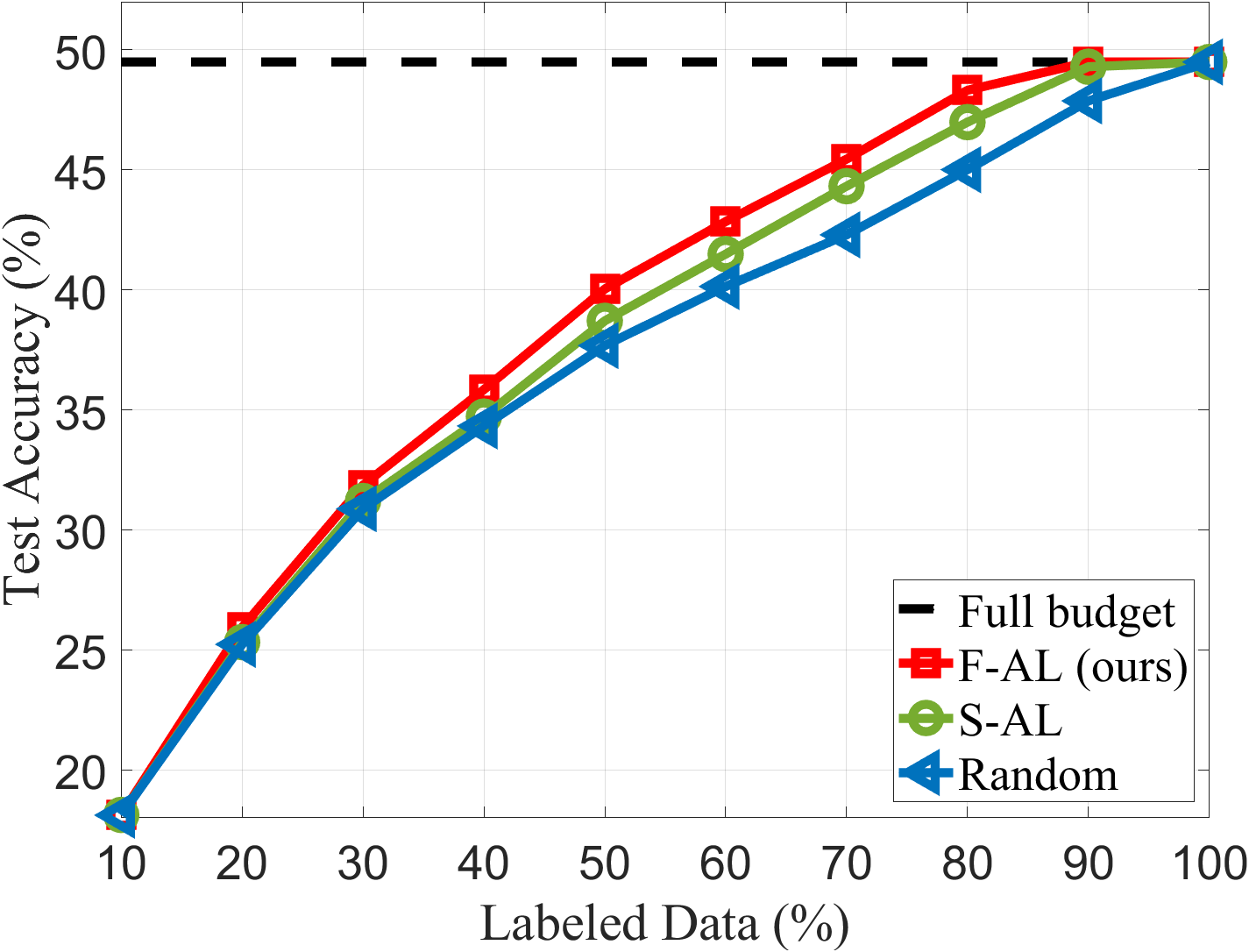}}
\caption{Test accuracies of global model trained by FL per rounds on CIFAR-100.}
\label{repre_cifar100}
\end{center}
\vskip -0.2in
\end{figure}

\subsection{Performance comparison}
Fig. \ref{repre_mnist}-\ref{repre_cifar100} illustrate the performance of random sampling (conventional FL), S-AL (benchmark), and F-AL (ours) which are the annotation strategies for FL. The AL algorithm is MCDAL, and the datasets are Fashion-MNIST, CIFAR-10, and CIFAR-100. Full budget in the figures denotes the performance of FL when all the clients have $100 \%$ labeled dataset. On the Fashion-MNIST, F-AL and S-AL considerably outperform random sampling, and the proposed F-AL shows the best performance compared to the other strategies. In particular, the average improvement compared to random sampling is $1.1 \%$ and $1.6 \%$ for S-AL and F-AL, respectively, at the 2nd round and 3rd round, before converging to the performance of the full budget. On the CIFAR-10, F-AL and S-AL outperform random sampling, and the proposed F-AL shows the best performance compared to the other strategies, same as the case of Fashion-MNIST. The average improvement compared to random sampling is $1.6 \%$ and $2.4 \%$ for S-AL and F-AL, respectively, before the 10th round. At the half of the rounds, the improvement is $2.3 \%$ and $3.9 \%$ for S-AL and F-AL, respectively when the performance of random sampling is $79.5 \%$.

On the CIFAR-100, it is also observed that the F-AL and S-AL show better performance than the performance of random sampling. The average improvement compared to random sampling is $1.1 \%$ and $2.0 \%$ for S-AL and F-AL, respectively, before the 10th round and the improvements are $2.0 \%$ and $3.2 \%$ at the $7$th round while the test accuracy of random sampling is $42.3 \%$. Fig. \ref{repre_mnist}-\ref{repre_cifar100} demonstrate that the proposed F-AL outperforms the baseline methods in the image classification of Fashion-MNIST, CIFAR-10, and CIFAR-100.    

\subsection{Extended results for various AL algorithms} 
\begin{figure}[t]
\vskip 0.2in
\begin{center}
\centerline{\includegraphics[width=0.4\columnwidth]{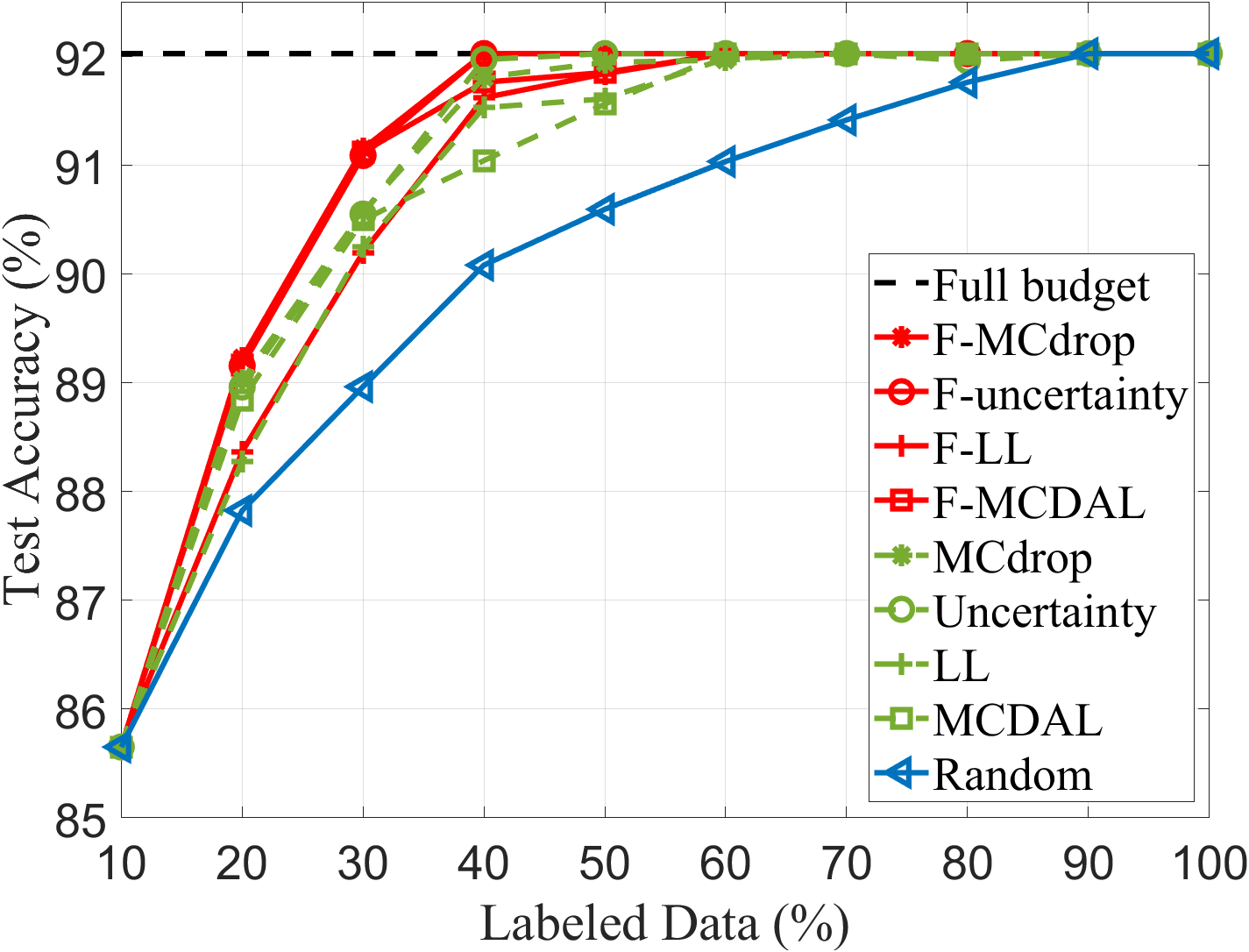}}
\caption{Test accuracies of global model trained by FL per rounds on Fashion-MNIST (uncertainty-related AL algorithms).}
\label{several_mnist}
\end{center}
\vskip -0.2in
\end{figure}

\begin{figure}[t]
\vskip 0.2in
\begin{center}
\centerline{\includegraphics[width=0.4\columnwidth]{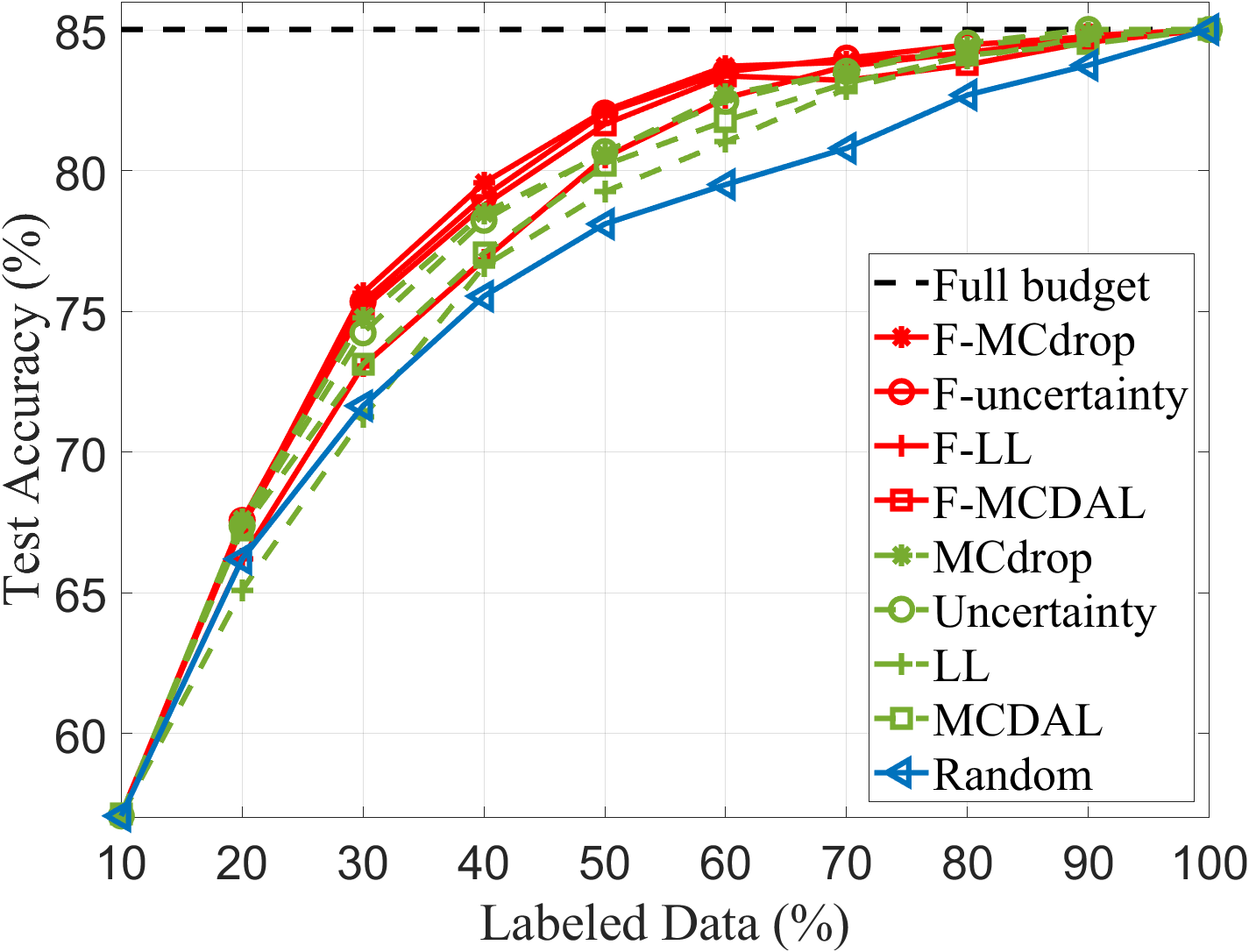}}
\caption{Test accuracies of global model trained by FL per rounds on CIFAR-10 (uncertainty-related AL algorithms).}
\label{several_cifar10}
\end{center}
\vskip -0.2in
\end{figure}

\begin{figure}[t]
\vskip 0.2in
\begin{center}
\centerline{\includegraphics[width=0.4\columnwidth]{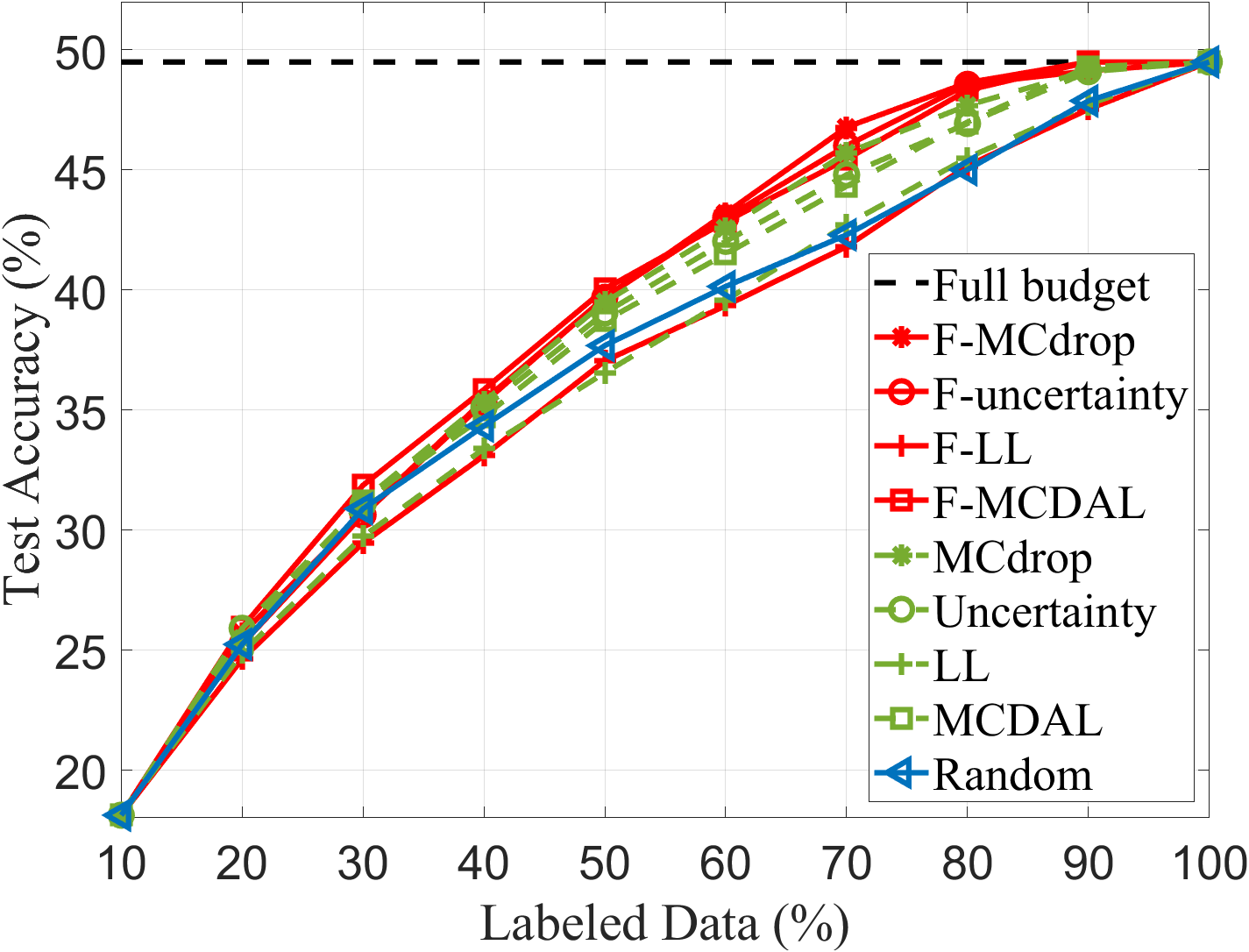}}
\caption{Test accuracies of global model trained by FL per rounds on CIFAR-100 (uncertainty-related AL algorithms).}
\label{several_cifar100}
\end{center}
\vskip -0.2in
\end{figure}

In order to demonstrate that our proposed F-AL outperforms the baseline methods for the general AL algorithms, we extend the experiment with MCDAL in Fig. \ref{several_mnist}-\ref{several_cifar100}. We first consider uncertainty-related AL algorithms: uncertainty-based sampling, MC-dropout with maximum entropy, LL, and MCDAL. Fig. \ref{several_mnist}-\ref{several_cifar100} illustrate that F-AL outperforms S-AL and random sapling for the considered AL algorithms. The only conflicting case is when LL is applied on the CIFAR-100, as observed in Fig. \ref{several_cifar100}. 

Through Fig. \ref{several_mnist}-\ref{several_cifar100}, we can compare the performance of the AL algorithms when they are applied for the FL environment. In Fig. \ref{several_mnist}, uncertainty-based sampling and MC-dropout show comparable performance, better than MCDAL and LL in both cases of S-AL and F-AL. In Fig. \ref{several_cifar10}, uncertainty-based sampling and MC-dropout show the best performance, and LL shows poor performance compared to the other algorithms, similar to the result in Fashion-MNIST. Fig. \ref{several_cifar100} illustrates that uncertainty-based sampling and MC-dropout outperform random sampling while MCDAL and LL show comparable performance with random sampling. As we observed in Fig. \ref{repre_cifar100}, F-MCDAL outperforms random sampling, but F-AL does not show considerable improvement for LL. In Table \ref{others_cifar10}, we consider the other categories of AL algorithms, which are the representative-based AL and adversarial AL. We evaluate the performance of random sampling, S-AL, and F-AL on CIFAR-10 when the AL algorithms are core-set approach and VAAL. Table \ref{others_cifar10} shows that both of core-set approach and VAAL show poor performance and are even worse than random sampling, while F-VAAL shows comparable performance with random sampling. F-AL does not show improvement when the core-set approach is applied for the AL algorithm.        

\begin{table*}[t]
\caption{Test accuracies of global model trained by FL per rounds on CIFAR-10 (core-set approach and VAAL).}
\label{others_cifar10}
\vskip 0.15in
\begin{center}
\begin{small}
\begin{sc}
\begin{tabular}{lccccccccccr}
\toprule
 \phantom{s} &\multicolumn{10}{c}{Labeled Data ($\%$)}   \\
methods    & $10\%$ & $20\%$ & $30\%$ & $40\%$ & $50\%$ & $60\%$ & $70\%$ & $80\%$ & $90\%$ & $100\%$ \\
\midrule
F-VAAL     &$0.571$ &$0.660$ &$0.720$ &$0.757$ &$0.778$ &$0.794$ &$0.808$ &$0.827$ &$0.839$ &$0.850$\\
VAAL     &$0.571$ &$0.634$ &$0.689$ &$0.737$ &$0.763$ &$0.789$ &$0.808$ &$0.819$ &$0.840$ &$0.850$\\
F-coreset     &$0.571$ &$0.652$ &$0.699$ &$0.742$ &$0.768$ &$0.791$ &$0.807$ &$0.820$ &$0.838$ &$0.850$\\
Coreset    &$0.571$ &$0.655$ &$0.700$ &$0.745$ &$0.770$ &$0.792$ &$0.806$ &$0.819$ &$0.838$ &$0.850$\\
Random     &$0.571$ &$0.662$ &$0.716$ &$0.755$ &$0.781$ &$0.795$ &$0.808$ &$0.827$ &$0.837$ &$0.850$\\
\bottomrule
\end{tabular}
\end{sc}
\end{small}
\end{center}
\vskip -0.1in
\end{table*}


\subsection{Discussion} In Fig. \ref{several_mnist}-\ref{several_cifar100} and Table \ref{others_cifar10}, it was first observed that uncertainty-based sampling and MC-dropout, which directly utilizes the uncertainty, show the best performance across most rounds of AL, and they have the largest performance increase by F-AL. In the previous literature \cite{yoo2019learning,cho2021mcdal}, it is validated that the LL and MCDAL outperform the classic uncertainty-based sampling and MC-dropout, contrary to the results in our experiments. In fact, LL and MCDAL learn the classifiers for discrepancy and the loss prediction module, respectively, in addition to the main task model, using the \textit{unlabeled dataset}. Compared to the large-scale dataset stored at one client in the literature \cite{yoo2019learning,cho2021mcdal}, multiple clients relatively have a much less number of instances in the unlabeled dataset for the distributed setting, e.g., $ 5$ clients respectively have $20 \%$ of the total dataset in our experiments. This insufficiency of the unlabeled dataset in the FL environment causes the worse performance degradation compared to the classical uncertainty-based AL algorithms.

With VAAL, the sampling step is implemented only by the variational autoencoder (VAE) and discriminator, trained by both the labeled dataset and the unlabeled dataset. Therefore, the performance is degraded by the insufficient unlabeled dataset, compared to the performance evaluated on the large-scale datasets \cite{sinha2019variational}. It is remarkable that F-VAAL outperforms VAAL, as observed in 
TABLE \ref{others_cifar10}, since the improvement comes from the development of VAE and discriminator with F-AL. The core-set approach generally shows excellent performance under the large-scale dataset \cite{sener2017active} since the representativeness-based AL algorithm alleviates the problem of uncertainty-based sampling which selects the similar instances near the decision boundary. However, the problem is suppressed in the distributed setting and the core-set approach performs poor in our FL environment. With the core-set approach, only the feature extraction of main task model is utilized for sampling, so that the core-set approach is rarely improved by the main task model, developed by F-AL. Furthermore, core-set approach requires whole datasets among clients to be stored at one site for the collaborative sampling in \eqref{samplingideal}. It means that the collaboratively sampled instances with the core-set approach should be
\be 
   \mathcal{L}^{k}=\underset{\mathcal{L} \subseteq \mathcal{U}^{k},~\left| \mathcal{L} \right|=\frac{b}{K}}{\mathrm{argmax}} ~ S\left(\mathcal{L}\; \middle| \; \bm{\phi}_{FL}^{k}  \right), 
\ee
but it cannot be obtained from the local computation as the \eqref{samplinginFAL2}, since it requires $\mathcal{U}^{k}$ in contrast to the FL constraint. Hence, F-AL cannot improve the performance of the core-set approach.

\begin{figure}[t]
\vskip 0.2in
\begin{center}
\centerline{\includegraphics[width=0.4\columnwidth]{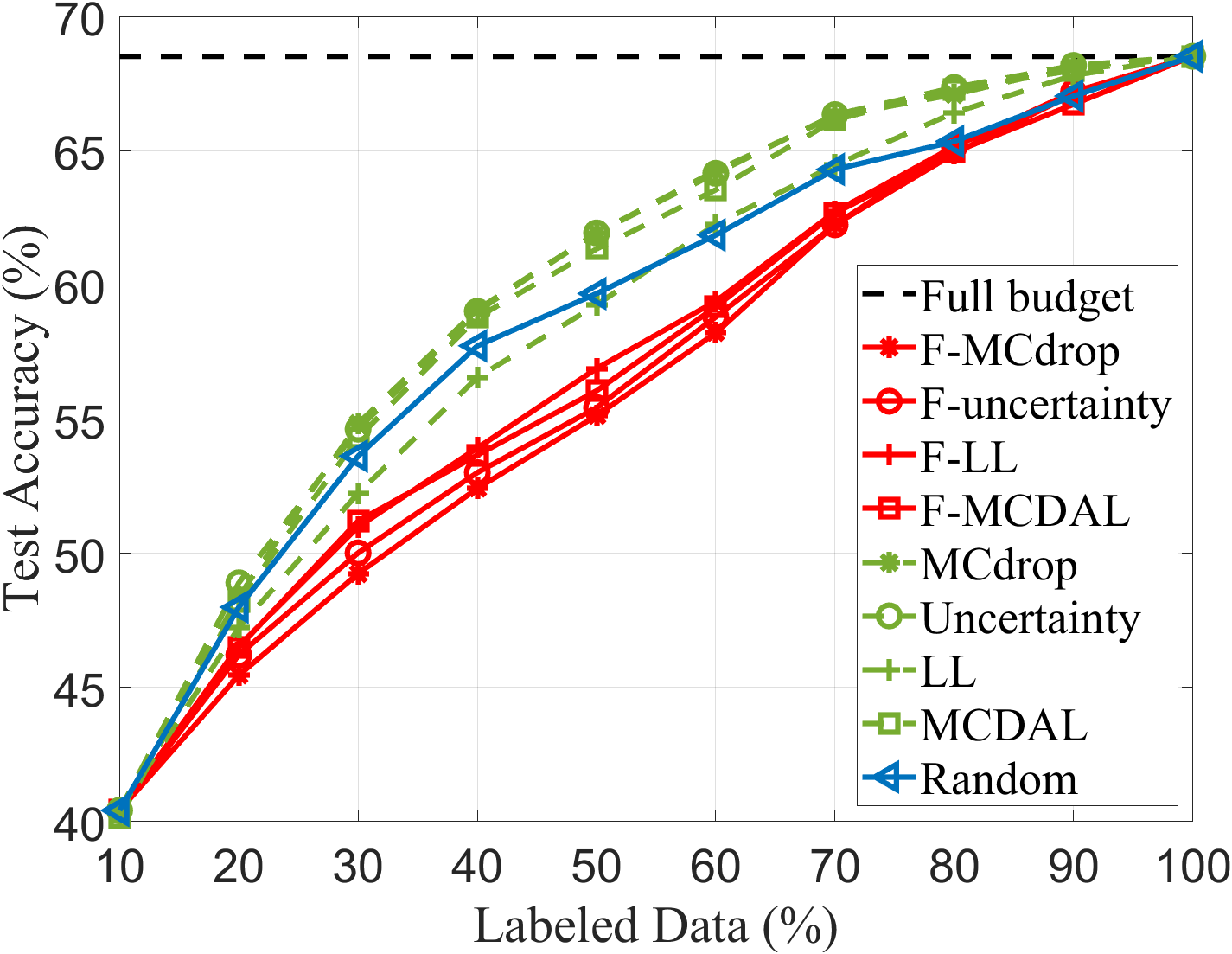}}
\caption{Test accuracies of model trained by independent learning per rounds on CIFAR-10.}
\label{IL_cifar10}
\end{center}
\vskip -0.2in
\end{figure}

\begin{figure}[t]
\vskip 0.2in
\begin{center}
\centerline{\includegraphics[width=0.4\columnwidth]{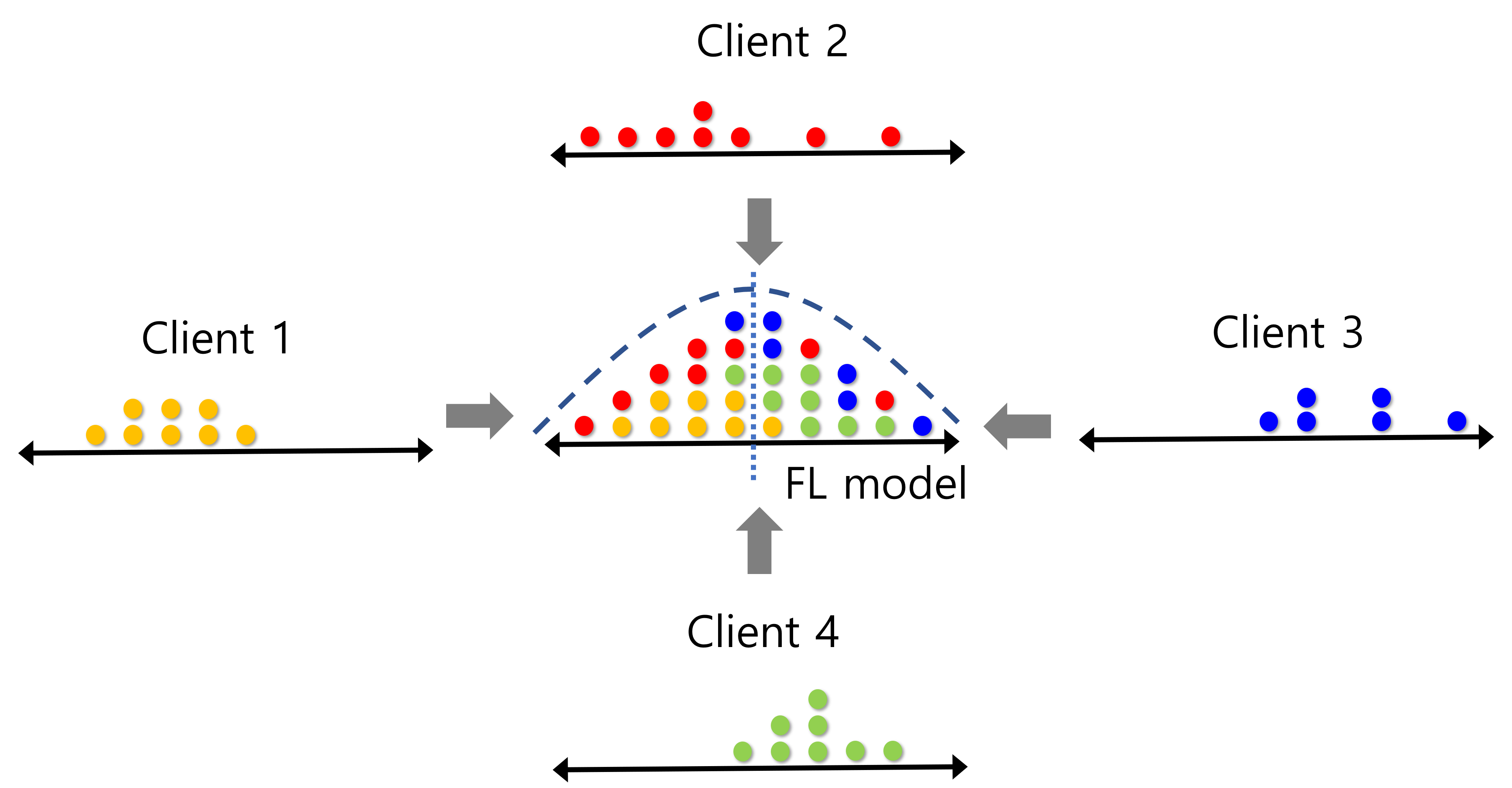}}
\caption{Distributions of the sampled instances }
\label{example}
\end{center}
\vskip -0.2in
\end{figure}

\textbf{Performance of independent learning} 
Through Fig. \ref{repre_mnist}-\ref{several_cifar100} and Table \ref{others_cifar10}, it has been demonstrated that AL is effective in FL environment, and the proposed F-AL outperforms the conventional random sampling and S-AL. For more discussion, we investigate the effect of F-AL in the perspective of local dataset. For this, each client solely trains the main task model with the local dataset after achieving the labeled dataset via the AL strategies. Fig. \ref{IL_cifar10} illustrates the average test accuracy of the models trained at the clients on CIFAR-10 when the AL algorithms are the uncertainty-related AL algorithms which show relatively significant performance increase by F-AL compared to the core-set approach and VAAL. It is observed that F-AL considerably decreases the performance of IL. In contrast, the S-AL certainly outperforms random sampling since S-AL samples the informative instances to the current local dataset. With F-AL, the clients collaborate to sample the informative instances to the aggregate datasets, not the local dataset. It becomes a solid constraint to the sampling of clients in the perspective of local datasets since each client with F-AL does not sample the instances that are not informative to the aggregate dataset even though the instances are informative to its datasets. As illustrated in Fig. \ref{example}, the aggregate dataset, which is sampled by F-AL, performs excellent for FL even though the sampled instances can be biased at the distribution of local datasets.

%% file: f.tex
\section{Conclusion}
In this paper, we focused on the active learning (AL) and sampling strategies into the FL framework to reduce the annotation workload. In our proposed federated active learning (F-AL) method, the clients collaboratively perform the AL to obtain the instances that can maximally improve the global model of FL. We empirically demonstrate that F-AL outperforms conventional random sampling strategy, client-level separate AL (S-AL) for the various AL algorithms on the image classification applications such as Fashion-MNIST, CIFAR-10, and CIFAR-100.